\pdfoutput=1

\documentclass[11pt]{article}

\usepackage{acl}

\usepackage{times}
\usepackage{latexsym}

\usepackage[T1]{fontenc}

\usepackage[utf8]{inputenc}

\usepackage{microtype}

\usepackage{graphicx}

\usepackage{marvosym}

\newcommand\blfootnote[1]{%
  \begingroup
  \renewcommand\thefootnote{}\footnote{#1}%
  \addtocounter{footnote}{-1}%
  \endgroup
}

\title{longhorns at DADC 2022: How many linguists does it take to fool a Question Answering model? A systematic approach to adversarial attacks.}

\author{Venelin Kovatchev\textsuperscript{1}\Cross* Trina Chatterjee\textsuperscript{3}* Venkata S Govindarajan\textsuperscript{2}* \\
\textbf{Jifan Chen\textsuperscript{3} Eunsol Choi\textsuperscript{3} Gabriella Chronis\textsuperscript{2} Anubrata Das\textsuperscript{1} Katrin Erk\textsuperscript{2}} \\
\textbf{Matthew Lease\textsuperscript{1} Junyi Jessy Li\textsuperscript{2} Yating Wu\textsuperscript{4} Kyle Mahowald\textsuperscript{2}*} \\
\textbf{*} Contributors towards the official submission \\
\textsuperscript{1} School of Information, The University of Texas at Austin \\
\textsuperscript{2} Department of Linguistics, The University of Texas at Austin \\
\textsuperscript{3} Department of Computer Science, The University of Texas at Austin\\ 
\textsuperscript{4} Department of Electrical and Computer Engineering, The University of Texas at Austin \\
 \\ 
}

\begin{document}
\maketitle
\begin{abstract}
Developing methods to adversarially challenge NLP systems is a promising avenue for improving both model performance and interpretability.
Here, we describe the approach of the team ``longhorns'' on Task 1 of the The First Workshop on Dynamic Adversarial Data Collection (DADC), which asked teams to manually fool a model on an Extractive Question Answering task.\blfootnote{\Cross Primary author and coordinator (\href{mailto:venelin@utexas.edu}{venelin@utexas.edu})}
Our team finished first, with a model error rate of 62\%.\footnote{The results and the team ranking are pending validation from the organizers of the task at the time of the submission.} 
We advocate for a systematic, linguistically informed approach to formulating adversarial questions, and we describe the results of our pilot experiments, as well as our official submission.
\end{abstract}

\section{Introduction}

Rapid progress in NLP has resulted in systems obtaining apparently super-human performance on popular benchmarks such as GLUE \cite{wang-etal-2018-glue}, SQUaD \cite{rajpurkar-etal-2016-squad}, and SNLI \cite{bowman2015large}. Dynabench \cite{kiela-etal-2021-dynabench} proposes an alternative approach to benchmarking: a dynamic benchmark wherein a human adversary creates examples that can ``fool'' a state-of-the-art model but not a  human language user.
The idea is that, by generating and compiling examples that fool a particular system, the community can gain a better idea of that system's actual strengths and weaknesses, as well as ideas and data for iteratively improving it.

There is no straightforward recipe, however, for generating successful adversarial examples.
To contribute to that knowledge base, this paper describes the  strategy used by team ``longhorns'' in Task 1 of The First Workshop on Dynamic Adversarial Data Collection (DADC), which was on Extractive Question Answering (answering a question about a passage by pointing to a particular span of text within that passage).\footnote{\url{https://dadcworkshop.github.io/shared-task/}} 
We focus not only on describing the details of our strategy, but also on our process for approaching the task. At the time of this paper submission, pending expert validation of the results, our team ranked first in the competition, obtaining 62\% Model Error Rate (MER). 

Our approach towards creating adversarial examples was designed to be systematic, analytical, and draw on linguistically informed ideas. 
We first compiled a list of linguistically inspired ``attack strategies'' and used it to create adversarial examples in a systematic manner.
We then analyzed some existing biases of the model-in-the-loop and its performance on a variety of different attacks. 
We used this piloting phase to select the best performing attacks for the official submission.

Based on the approaches that were most successful both in our pilot studies and in our official submission, we posit that the following broad areas should be of particular interest for theoretically motivated adversarial attacks on contemporary NLP systems, as evidenced by their strong performance on our target task:

\begin{itemize}

\item 
\textbf{Taking advantage of models' strong priors.}
The model was proficient at identifying the correct kind of named entity being asked for (e.g., a person for a "who" question, a place for a "where" question), but was biased to give answers which were salient (either topically or because they appeared first~\cite{Ko2020LookAT}) or which had high lexical overlap with the question. Thus, picking a distractor with the same entity type as the target answer (e.g., another person mentioned in the text when the question was a ``who" question) was often effective. This result is broadly consistent with observations that modern NLP systems can perform well in the general case but can be biased towards frequency-based priors \citep[e.g.,][]{wei-etal-2021-frequency} that mean they are sometimes  ``right for the wrong reasons'' \citep{mccoy-etal-2019-right}.

\item 
\textbf{Using language that is linguistically taxing for humans (and machines) to process.} 
Psycholinguists who study human language processing often study constructions that are grammatical but difficult for humans to process in real time, such as garden path sentences \citep{frazier1982making,ferreira1991recovery} and complex coreference resolution \citep{kaiser2019reference,durrett-klein-2013-easy}.
We found that the model was indeed often fooled by questions that included these types of constructions.
While we did not collect any human data, the sentences that fooled the model are likely to be hard for humans as measured by tests of real-time processing difficulty (e.g., eye tracking, self-paced reading), even though humans would be able to successfully process these sentences given enough time.

\item 
\textbf{Tapping into domain-general, non-linguistic reasoning.} 
We found that asking questions which do not require mere linguistic processing but require other kinds of reasoning (e.g., numerical reasoning, temporal reasoning, common-sense reasoning, list manipulation) were hard for the model.
This result is consistent with prior work showing that language models struggle with these kinds of reasoning tasks \citep{marcus2020next,elazar-etal-2021-back,talmor2020olmpics} and may be more generally explained by evidence from cognitive science that these kinds of reasoning tap into cognitive processes that are distinct from linguistic processing \citep{diachek2020domain,blank2014functional}. 

\end{itemize}

Because these strategies and this general approach are broad and theoretically motivated, we believe that our methods could be used to generate adversarial examples on other Natural Language Understanding tasks besides Question Answering. 
In what follows, we characterize our approach in both the pilot phase and official submission, provide our list of attack strategies, 
and discuss the limitations of the task and model.

\section{Task Definition}

In Task 1 of DADC, titled ``Better Annotators'', each participating team submits 100 ``official'' extractive question answering (QA) examples through the Dynabench platform.
The organizers of the shared task provide short passages as context and the participants have to create questions that ``can be correctly answered from a span in the passage and DO NOT require a Yes or No answer''. 
The objective is to find as many model-fooling examples as possible -- the winning team is the one with the highest validated model error rate (vMER)\footnote{For full instructions, see \url{ https://dadcworkshop.github.io/shared-task/}}.

The competition uses Dynabench \cite{kiela-etal-2021-dynabench}: ``an open-source platform for dynamic dataset creation and model benchmarking''. Dynabench aims to facilitate human-and-model-in-the-loop dataset creation. The annotators' aim to generate examples that will be misclassified by an automated model, but can be answered correctly by competent human speakers. Dynabench has been used to create data for Question Answering \cite{kaushik-etal-2021-efficacy}, Natural Language Inference \cite{williams-etal-2022-anlizing}, Online Hate Detection \cite{vidgen-etal-2021-learning}, and Sentiment Analysis \cite{potts-etal-2021-dynasent}, among others.

\section{Approach}\label{dadc:apr}

Our team consisted of faculty, postdocs, and students from the UT Austin linguistics department, computer science department, information school, and electrical and computer engineering department.

We approached the problem of creating adversarial attacks in a systematic manner, informed by ideas from computational linguistics, psycholinguistics, and theoretical linguistics.
We composed a list of linguistic 
phenomena and reasoning capabilities that we hypothesized would be difficult for a state-of-the-art QA model. 
We then used some of those phenomena to create our official submission of adversarial examples. 
While the list is not exhaustive, it covers a wide range of potential attack strategies and can be used to guide the creation of adversarial examples for other tasks and systems. 

Separate from our official submission for the competition, we ran a series of pilot experiments in which we used the list as a guide for experimenting with a variety of strategies for creating adversarial example.
To ensure a fair and competitive official submission, all pilot experiments were carried out either before the official start of the shared task or after the official submission was made. 


Our objective when evaluating the different adversarial strategies was to explore the space of potential attack strategies to determine the most successful ones for fooling the model. For each strategy, we measured the Model Error Rate (MER) on a small sample of example, and we also analyzed how frequently the attack can be used. 


Based on the results of these pilot experiments, we targeted the best strategies for our official submission.
Official question submissions were made by subsets of the team, in group sizes ranging from 1 to around 10. 
Since not all strategies can be used for all example passages, we used the specific passages we were presented with in order to guide our decision about what strategy to focus on for a particular question.
When more than one participant was present, question submissions were made by consensus agreement among those present. 

Anecdotally, we found that the attacks were often more successful when multiple team members are present, with each member hypothesizing model behaviors from diverse angles. Overall, we found the adversarial question generation process nontrivial, taking 5-10 minutes per passage, although we became faster over time. We also chose to skip passages occasionally when the passage covers very well-known entity, is too simple, or is not promising to most of our strategies (not having distractor entities, etc). We generated multiple questions for promising passages.

\section{Pilot Experiments: Evaluating Adversarial Strategies}\label{dadc:sandbox}

Many of our ``adversarial strategies'' are inspired by prior work in adversarial data generation and unit testing for Question Answering, Natural Language Inference, and Paraphrase Identification \cite{glockner-etal-2018-breaking,kovatchev-etal-2018-etpc,naik-etal-2018-stress,dua-etal-2019-drop,kovatchev-etal-2019-qualitative,Nie_Wang_Bansal_2019,wallace-etal-2019-trick,DBLP:journals/corr/abs-2002-00293,Gardner2020-zw,hossain-etal-2020-analysis,jeretic-etal-2020-natural,Kaushik2020Learning,ribeiro-etal-2020-beyond,saha-etal-2020-conjnli}. We propose the following linguistic and reasoning phenomena as a source for potential adversarial attacks:

\paragraph{Lexical knowledge}

Examples that require understanding lexical properties and in partiuclar lexical entailments that require knowledge of \textbf{hypernyms/hyponyms} (e.g., knowing that dog => animal, but animal =/> dog), \textbf{named entities} and their properties (e.g., knowing Shakira is a singer), \textbf{nominalization} (e.g., knowing ``a submission'' implies something has been submitted), \textbf{(a)symmetrical relations} (e.g., knowing that John marrying Mary implies Mary marrying John, but John loving Mary does not imply Mary loving John), \textbf{polarity substitutions} (e.g., knowing that a movie is good implies that it is not bad), \textbf{converse substitution} (e.g., knowing that if something has been provided, it has been received), 
\textbf{comparisons with antonyms} (e.g., knowing that if Clara is the tallest, she is not shorter than Mary), \textbf{reasoning about modal verbs} (e.g., understanding that if something \textit{could} happen, that does not mean it \textit{did} happen), and \textbf{reasoning about quantifers} (e.g., knowing that if some swans are white, that does not imply all swans are white).

\begin{figure*}[h!]
    \centering
    \includegraphics[width=0.9\textwidth]{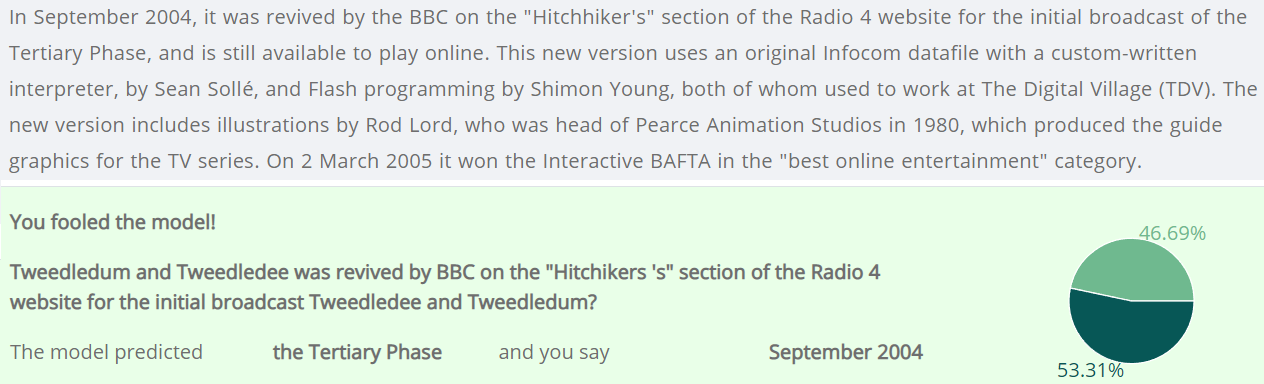}
    \caption{An example of ``semantic similarity'' model bias: when the model is fed a nonsensical question, it responds with an answer with high semantic overlap with the question. }
    \label{fig:surf}
\end{figure*}

\paragraph{Syntax and Discourse knowledge} 

Examples that require syntactic or discourse-level understanding such as 
\textbf{Genitives} (e.g., knowing that elephant's foot = the foot of the elephant) and 
\textbf{Datives} (e.g., knowing that give her a cake = give a cake to her). 
\textbf{Relative Clauses} can be used in attacks to either include distracting information (e.g., ``Maria, who is the president of the company'' when the correct answer has nothing to do with Maria's role in the company) or to specify the correct referent (e.g., ``the actor who bought the house'' when that actor must be distinguished from a set of other actors).

When \textbf{Conjunction} or \textbf{Disjunction} appear in the passage (e.g., John and Mary love strawberries and cake, but John doesn't like chocolate), an adversarial question targets the ability of the model to correctly identify the syntactic scope (e.g.: Who loves cake and chocolate?). 
Closely related are the phenomena of \textbf{Intersectivity} (e.g., knowing that ``a singer and a good man'' =/> a good singer) and \textbf{Restrictivity} (e.g., understanding that ``all my work due today'' =/> all my work).

When a complex \textbf{prepositional phrase attachment} appears in the passage (e.g., ``I saw two men with a telescope in the park''), an adversarial question requires disambiguation (e.g., ``Who has the telescope''). 
Questions based on the \textbf{Argumentative structure} require the model to correctly identify the core arguments (e.g., knowing that ``John broke the vase'' implies that ``The vase broke''; but does not imply ``John broke.''). This attack can be further complicated when the same verb appears multiple times in the passage.
Adversarial attacks based on \textbf{Ellipsis}, \textbf{Anaphora}, and \textbf{Coreference} test the ability of the model to process long distance syntactic dependencies. 

\textbf{Negation} can appear both in the passage and in the question. It can be expressed in a variety of ways: simple (e.g., no, not), adverbial (e.g., never), pronoun (e.g., nobody), morphological (e.g., unfinished), lexical (e.g., refuse to), implicit (e.g., I wish I had a boat), double negation. Adversarial questions can also target the ability of the model to identify the scope of negation either in the question or in the passage.

\textbf{Garden Path questions} (e.g., Who is the director of the movie directing?) are syntactically confusing and much-studied in psycholinguistics for causing processing difficulty in humans \cite{frazier1979comprehending}.

Questions about \textbf{Mental States} of individuals are inspired by work in psychology showing that it can be challenging to reason about the mental states of others (e.g., knowing that ``Why does Maria think that Sandra is leaving?" could require a different answer than ``Why is Sandra leaving?''\cite{wellman1992child,kovatchev-etal-2020-mind}.

\paragraph{Reasoning} 

Questions that require various kinds of non-linguistic reasoning such as \textbf{Conditionals and hypothetical situations} (e.g., Who would be the champion if Mary didn't lose the final?), \textbf{Numerical Reasoning} (e.g., Who is the second richest person?), \textbf{Temporal Reasoning} (e.g., What happened in a specific timeframe?), \textbf{Commonsense reasoning} (including logical implications, contradiction, etc.), and \textbf{List manipulations} (e.g., Which two of the actors in the list are male?).

Finally, \textbf{distractor}-based attacks make use of model priors by expanding the question with additional information. \textbf{Meaningful distractors} directs the model towards a wrong answer, while the strategy of \textbf{adding noise} relies on increasing the complexity of the question.

The different phenomena can appear in the passages, in the question, or in both. Not every phenomena can be used to generate attacks for every passage, and the phenomena also appear with different frequency in the data. In our pilot experiments we distributed the different phenomena across the members of the team. We measured the Model Error Rate (MER) for the different strategies and determined how frequently each attack could be used.

\begin{figure*}[h!]
    \centering
    \includegraphics[width=0.9\textwidth]{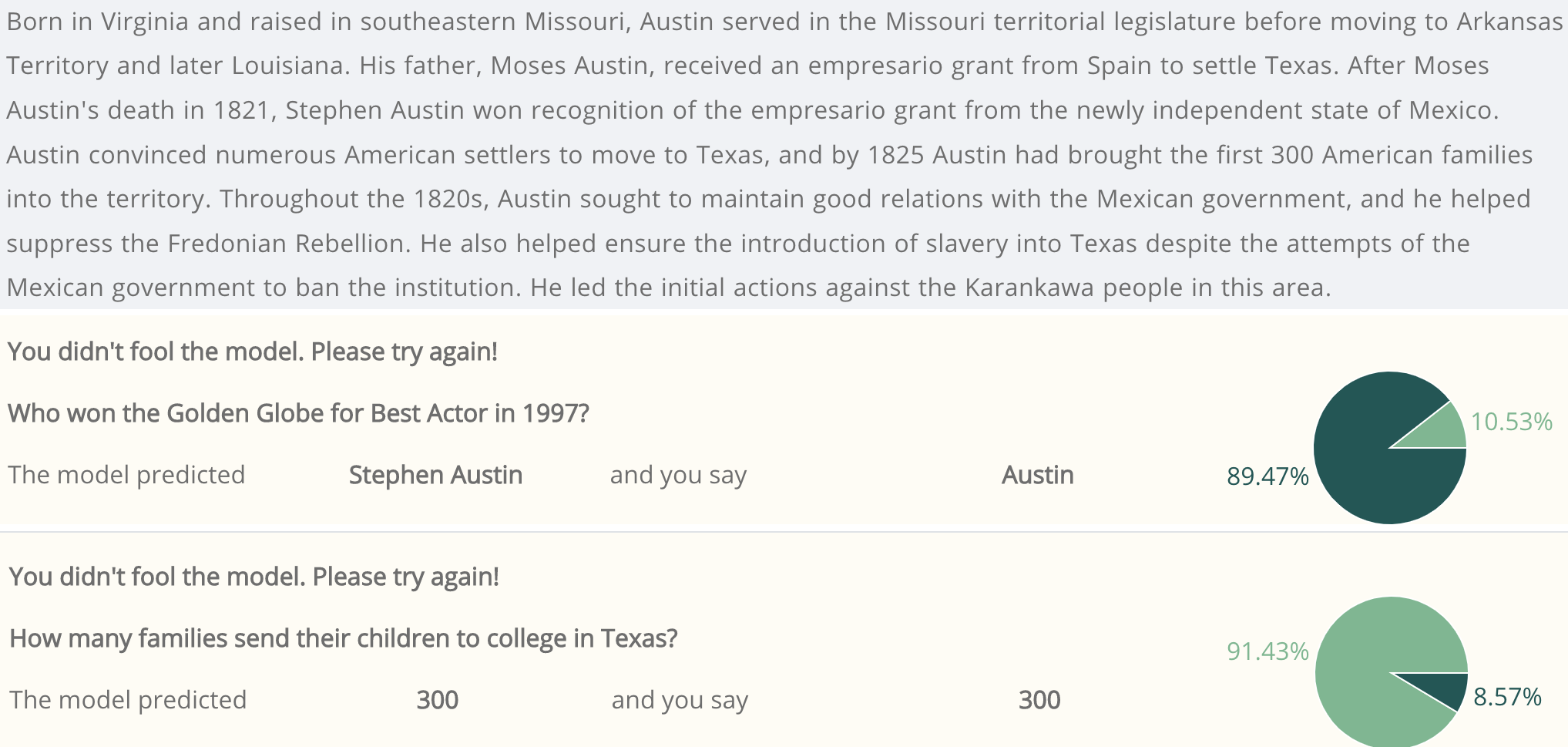}
    \caption{When faced with nonsensical questions, the model will give salient answers of the correct entity type. The passage is about Texas colonizer Stephen Austin, but the first question is about the Golden Globe awards. The model confidently answers that Stephen Austin won the Gold Globe for Best Actor in 1997.}
    \label{fig:type}
\end{figure*}

\section{Exploring Model Biases}

During the pilot experiment step in Section \ref{dadc:sandbox} we found that the model-in-the-loop performs surprisingly well on a variety of different attacks. We hypothesized that at least in some situations, the strong performance is due to spurious correlations, the nature of the underlying language model, and the nature of the task. We further carried out a set of experiments to determine some specifics of the model behavior. We briefly discuss two "shortcuts" used by the model.

\paragraph{Semantic similarity}

Figure \ref{fig:surf} illustrates the model bias towards ``semantic similarity'' on a nonsensical question. When the model is unsure what to do, or like in this example, when the question is not a valid English sentence, it identifies parts of the context that are similar to the question and predicts neighboring words. 
Due to the relatively short length of most of the passages, this strategy often gets the correct answer without actually understanding the question.

\paragraph{Type of question}

We noticed that the model is very good at identifying some properties of the answer based on the type of question. For example, a ``who'' question typically asks for a named entity, while a ``how many'' question asks for a quantity. A strong heuristic adopted by the model is to return an answer of the correct ``type'' regardless of the actual question. Figure \ref{fig:type} illustrates that: neither question can be answered from the passage, but the model makes a guess based on the type of question. Once again, the short length of the passages and the fact that they typically contain just a few tokens of the correct ``type'' artificially boosts the performance of the model.

In our official submission, we used those model biases to increase the difficulty of the adversarial examples. When possible, we used those biases to guide the model towards a wrong answer. In passages where we could not confuse the model (e.g.: only one named entity in a ``who'' question), we rephrased the questions in such a way that makes it harder for the model to use heuristics.

\begin{figure*}[h!]
    \centering
    \includegraphics[width=0.9\textwidth]{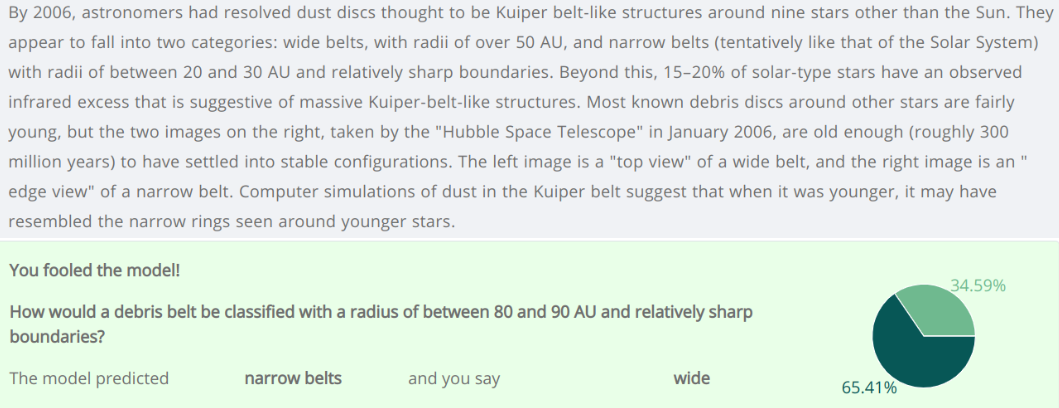}
    \caption{An example of ``distractor'' and ``numerical reasoning'' strategies. The model has to reason that ``between 80 and 90'' is greater than the 50 AU boundary identified in the passage.}
    \label{fig:distr}
\end{figure*}

\begin{figure*}[h!]
    \centering
    \includegraphics[width=0.9\textwidth]{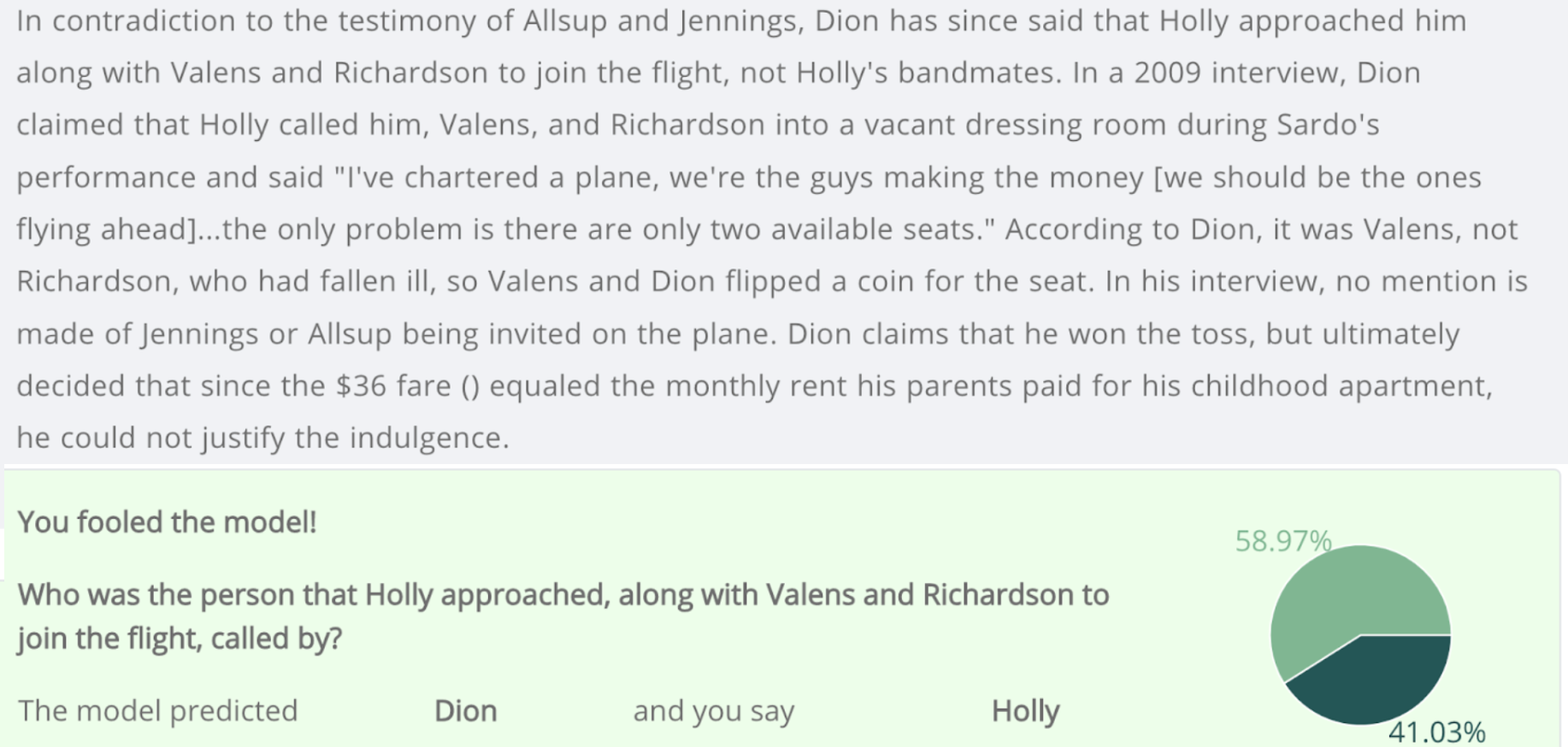}
    \caption{``Garden path'' strategy. Until the very end of the sentence, the question seems to be about something else.}
    \label{fig:gard}
\end{figure*}

\section{Official Submission}

After analyzing and discussing the results of our preliminary experiments, for our official submission we focus on the following strategies: using distractors, numerical reasoning, temporal reasoning, garden path questions, complex coreference, list manipulations, and common-sense reasoning. 
We also used the model biases to either confuse the model or reduce it's ability to rely on heuristics. 
In the rest of this section we briefly describe each of our strategies and provide examples.

\subsection{Taking advantage of model priors}

\paragraph{Distractors}

One of the most successful and easy-to-use adversarial strategies was using distractors. An example of that strategy can be seen in Figure \ref{fig:distr}: the phrasing of question has a high degree of similarity with the portion of the text talking about narrow belts (``between X and Y AU and relatively sharp boundaries''), however the correct response is ``wide belts'' due to the specified sizes. The ``distractor'' strategy can be used frequently. We often combined the distractor strategies with other strategies. For example, in Figure \ref{fig:distr}, we combine it with ``numerical reasoning''. 
Anecdotally, we found the distractors to be most successful when the correct answer was not the most salient entity of its type in the passage (e.g., targeting a briefly mentioned director in a passage mostly about one particular actor) and when there were many other entities of the desired type available, as opposed to just 1 or 2 (e.g., a ``who" question for a passage that mentions 10 people is more challenging than a ``who" question for a passage that mentions only 1 person).


\subsection{Linguistically difficult utterances}

\paragraph{Garden Path Questions}

Figure \ref{fig:gard} shows an example of a garden path question \citep{frazier1979comprehending}.
Until the last word, the question appears to be asking about ``the person that Holly approached, along with Valens and Richardson'' (to which the answer would be Dion). But the last word makes it clear that the reader needs to reparse the question, to see that it is actually asking who \textit{calls} that person (i.e., Dion) -- which makes the answer Holly.
The model is unable to correctly process the complex syntactic structure of the sentence and responds ``Dion''.
Garden path questions are easier to generate than temporal and numerical reasoning questions since they can be generated for a wider variety of texts, but we found that the model can often handle even quite complex syntactic constructions. We hypothesize that this is mainly due to the length of the passages and the ``type of question'' model bias.

\begin{figure*}[h!]
    \centering
    \includegraphics[width=0.9\textwidth]{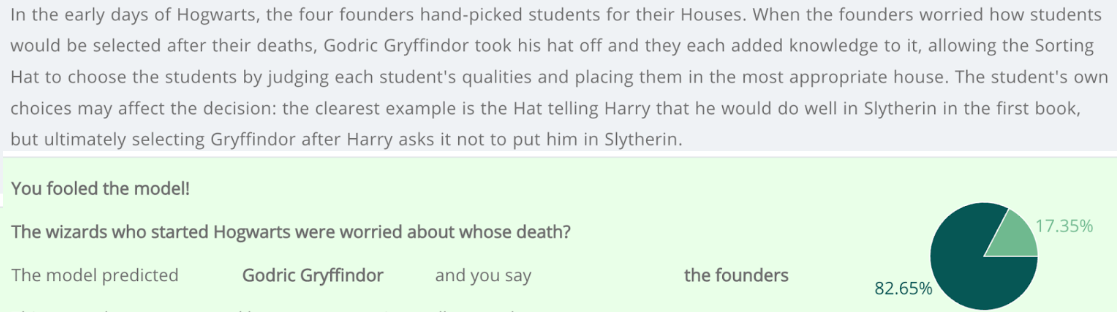}
    \caption{``Coreference'' strategy. The model has to figure out that the word ``their" refers to the founders.}
    \label{fig:cor}
\end{figure*}

\paragraph{Anaphora and Coreference}

Adversarial examples based on anaphora and coreference can require the model to demonstrate the ability to resolve long distance syntactic dependencies and often require making common-sense inferences as well.
In Figure \ref{fig:cor}, the founders were worried about their own death. 
To correctly respond to the question, the model first has to identify ``their'' as the answer and then resolve the coreference between ``their'' and ``the founders''. Instead, the model just returns a salient named entity. 
Examples based on anaphora and coreference are relatively infrequent, as they require multiple entities and potentially ambiguous coreference in the passage.

\begin{figure*}[h!]
    \centering
    \includegraphics[width=0.9\textwidth]{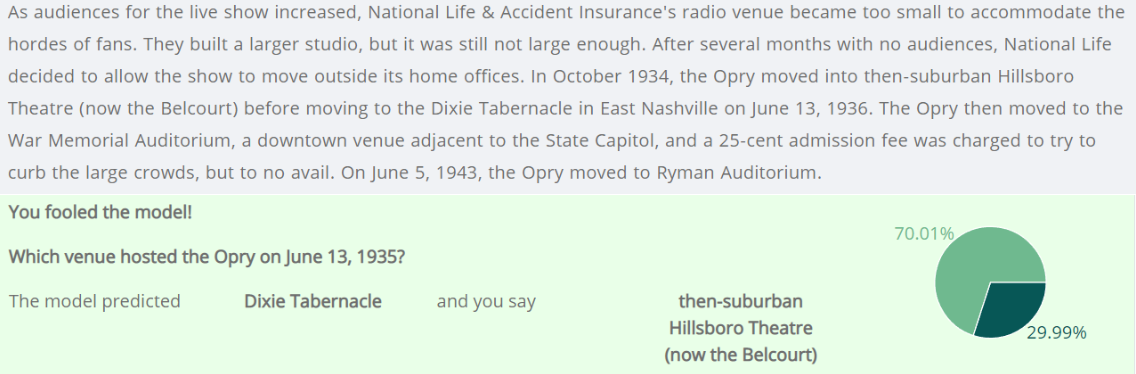}
    \caption{``Temporal reasoning'' strategy. The model has to understand that June 13, 1935 is during the period when the Opry moved to the Hillsboro Theatre.}
    \label{fig:temp}
\end{figure*}

\begin{figure*}[h!]
    \centering
    \includegraphics[width=0.9\textwidth]{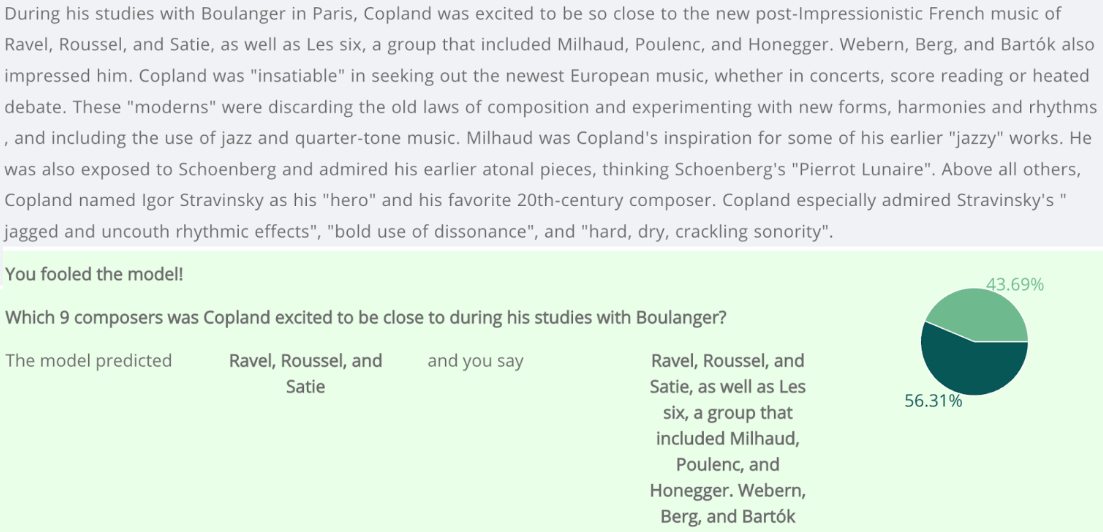}
    \caption{``List manipulations'' strategy. The model has to identify the 9 composers asked for, but only gives 3.}
    \label{fig:list}
\end{figure*}

\begin{figure*}[h!]
    \centering
    \includegraphics[width=0.9\textwidth]{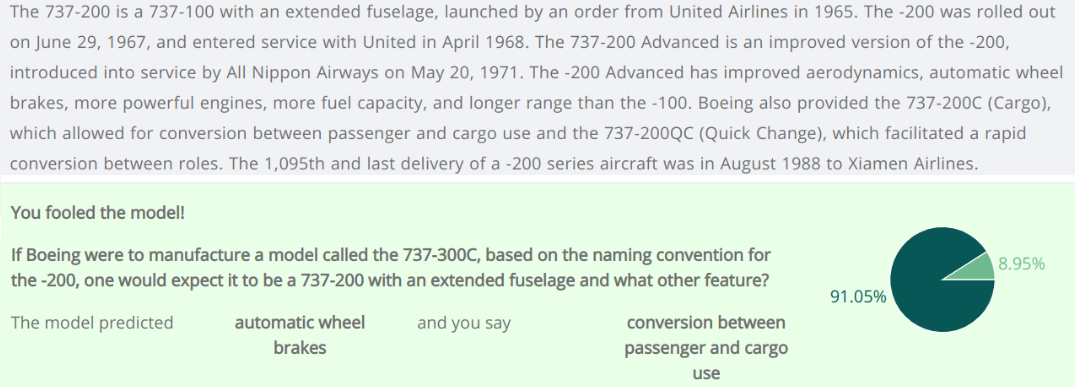}
    \caption{``Common-sense reasoning'' strategy. The model has to flexibly adapt the naming convention to a hypothetical example: a kind of creative reasoning task that humans do easily but that models often struggle with.}
    \label{fig:reason}
\end{figure*}

\subsection{Non-linguistic reasoning}

\paragraph{Numerical Reasoning}

Adversarial examples based on numerical reasoning require the model to carry out simple mathematical calculations or comparisons to identify the correct answer. For example, in Figure \ref{fig:distr}, the model had to calculate that ``between 80 and 90'' is ``over 50'' in order to answer correctly.

\paragraph{Temporal Reasoning}

Adversarial examples based on temporal reasoning require the model to reason about the chronological order of events and the different states of the world at different points in time. In the example shown in Figure \ref{fig:temp} the Opry moves to Hillsboro in 1934 and then to Dixie Tabernacle in 1936. We ask the model for a date that is not mentioned explicitly (June 13, 1935). The correct answer is ``Hillsboro'', but the model is fooled by recognizing a portion of the date (June 13) and predicts Dixie Tabernacle.

Temporal-based examples are relatively rare, as they require the passage to  have multiple dates as well as multiple different events and world states associated with the dates. However, when available, temporal-based attacks were very successful. 

\paragraph{List Manipulations}

We used two different strategies to create adversarial examples based on lists. Figure \ref{fig:list} illustrates one of them. The question requests the full list of 9 composers, while the model only extracts the first three due to the syntactic structure of the sentence.
The second list-based strategy asks for a subset of a list that fulfills certain criteria. List-based adversarial attacks are relatively infrequent in a single passage setting we study.

\paragraph{Common-sense Reasoning}

Adversarial attacks based on common sense reasoning test the basic understanding of the world of the model or its ability to reason about different entities and objects. In Figure \ref{fig:reason}, the model is required to break apart the string ``737-300C'' and ``737-200C'' correctly and then reason about the naming convention: we are told that ``C'' stands for cargo and so the hypothetical ``737-300C'' should also have the cargo feature.

\section{Discussion}

A fundamental feature of language is that it is a cooperative enterprise \citep{clark1996using} that enables efficient communication between parties \citep{gibson2019efficiency}.
Therefore, in ordinary language, people typically talk about discourse-relevant entities \citep{sperber1986relevance}, avoid difficult syntactic constructions \citep{futrell2015largescale,gibson1998linguistic}, and structure information in a way that is easy to produce and understand \citep{macdonald2013language,levy2008expectation}.
If anything unifies all of our most successful attack strategies, it is that they eschew these principles in the context of the given task and passages.
Instead, successful attacks ask about surprising aspects of the text (e.g., by including  distractors), often using complex language (e.g., garden path sentences and complex coreference resolution) and reasoning (e.g., temporal and numeric reasoning).


So, in some ways, the successful attack questions are less likely to be encountered in ordinary language use \citep[leading to claims that adversarial examples are brittle, e.g.,][]{phang2021adversarially,bowman-2022-dangers}. 
But another key property of human language is that it is flexible and generative, such that people can produce and understand surprising and unexpected utterances.
To that end, we think these adversarial questions are a fair target for improving  systems precisely because they are linguistically unusual: human language is not just for the ``average case" but can be used to express meanings that are subtle, interesting, and complicated. 

Perhaps because these questions also require humans to think creatively outside their ordinary linguistic experience, we also found that we achieved better performance when we had larger groups of people working on generating questions at once, so that there was a wider diversity of ideas.

Indeed, while some questions may be less likely to appear in a ``extractive question answering'' dataset, they are understandable by humans and are likely to be useful for efficient communication in real-world settings. The objective behind ``extractive QA'' is that a machine should answer any question that a human would, given the passage. A variety of real-world tasks can be reduced to extractive QA and in many cases the ``correct'' passage corresponding to the question is not known a priori. Asking questions such as ``Where was X at a time Y'' and ``What is the difference between 737-200 and 737-200C'' may be less natural for a human that has access to the passage, but are questions that someone would, for example, ask their automated assistant. Therefore, a well functioning model needs to embrace the creativity and be able to correctly answer adversarial questions.


Finally, the adversarial attacks that we present are not just interesting from the scientific point of view, but also have clear practical implications. Most of the attacks correspond to specific capacities of the model-in-the-loop such as coreference resolution, numerical and temporal reasoning.
The consistently high MER indicates that the model underperforms in tasks that require those capacities.



Our approach towards creating adversarial examples allows us to implicitly evaluate the performance of the model and the quality of the data with respect to a wide variety of linguistic and reasoning categories. Overall, we found that the model-in-the-loop performs impressively good on the majority of question types. Only a small subset of the strategies could consistently obtain above 50\% MER and these strategies did not necessarily work for all questions. 
For instance, questions with relatively few possible entities matching the question type meant fewer possibilities for distractors. 

The performance of the model is also a function of the varying difficulty of the passages. We found the majority of the passages to be short declarative texts with a simple syntactic structure, few named entities, and low amount of information. Generating and answering questions from those passages is a rather trivial task. The selection of more complex paragraphs will likely result in a lower performance of the model and a lot more possibilities for creative and successful adversarial attacks.

\section{Conclusions}

In this paper we presented the strategies used by team ``longhorns'' for Task 1 of DADC: generating high-quality adversarial examples. We obtain the best results in the competition by taking a systematic approach, using linguistic knowledge, and working in a collaborative environment.

Our approach outperforms prior work in terms of model error rate and also provides a variety of insights. For instance, our pilot analysis covers a large number of linguistic and reasoning phenomena and explores different model biases. This facilitates a more in-depth analysis of the performance of the model. The systematic approach also gives us insight into the quality and difficulty of the data.

Our strategies for generating adversarial examples are not limited to extractive question answering. They can be adopted at larger scale to improve the quality of models and data on a variety of different tasks. We believe that our work opens new research directions with both scientific and practical implications. 

\section*{Acknowledgements}

This research was supported in part by NSF grants IIS-1850153 and IIS-2107524, as well as by Wipro, the Knight Foundation, the Micron Foundation, and by Good Systems,\footnote{\url{http://goodsystems.utexas.edu/}} a UT Austin Grand Challenge to develop responsible AI technologies. The statements made herein are solely the opinions of the authors and do not reflect the views of the sponsoring agencies.

\bibliography{anthology,custom}
\bibliographystyle{acl_natbib}

\appendix

\end{document}